\documentclass[letterpaper, 10 pt, conference]{ieeeconf} 
\IEEEoverridecommandlockouts                              
\usepackage{graphics} 
\usepackage{epsfig} 
\usepackage{mathptmx} 
\usepackage{times} 
\usepackage{amsmath} 

\usepackage{amssymb}  
\usepackage{siunitx}
\usepackage[euler]{textgreek}
\usepackage{float}
\usepackage{cite}

\begin{document}
    \title{\Large \bf
    High-curvature, high-force, vine robot for inspection}
  \author{
      Mijaíl Jaén Mendoza$^{1*}$, Nicholas D. Naclerio$^{1}$, and Elliot W. Hawkes$^1$
      \thanks{This work was supported in part by the National Science Foundation under
Grant 1944816 and by a David and Lucile Packard Foundation Fellowship for Science 
and Engineering.}
  \thanks{$^1$Department of Mechanical Engineering, University of California, Santa Barbara, CA 93106.}
   \thanks{$^*$ Corresponding author. Email: mmendozaflores@ucsb.edu}
}


\maketitle

\begin{abstract}

Robot performance has advanced considerably both in and out of the factory, however in tightly constrained, unknown environments such as inside a jet engine or the human heart, current robots are less adept.
In such cases where a borescope or endoscope can't reach, disassembly or surgery are costly.
One promising inspection device inspired by plant growth are ``vine robots" that can navigate cluttered environments by extending from their tip.
Yet, these vine robots are currently limited in their ability to simultaneously steer into tight curvatures and apply substantial forces to the environment.
Here, we propose a plant-inspired method of steering by asymmetrically lengthening one side of the vine robot to enable high curvature and large force application. 
Our key development is the introduction of an extremely anisotropic, composite, wrinkled film with elastic moduli 400x different in orthogonal directions.
The film is used as the vine robot body, oriented such that it can stretch over 120\% axially, but only 3\% circumferentially.
With the addition of controlled layer jamming, this film enables a steering method inspired by plants in which the circumference of the robot is inextensible, but the sides can stretch to allow turns. 
This steering method and body pressure do not work against each other, allowing the robot to exhibit higher forces and tighter curvatures than previous vine robot architectures. 
This work advances the abilities of vine robots--and robots more generally--to not only access tightly constrained environments, but perform useful work once accessed.




\end{abstract}

\section{Introduction}

When a complicated mechanical system such as a jet engine, nuclear reactor, or human heart malfunctions it often requires an internal inspection to diagnose or solve the problem. 
Some simple problems can be accessed by removing an external component or using a borescope or endoscope to peer into the machine.
However, many internal problems require disassembly or surgery due to the inadequacy of current inspection tools, a costly and time intensive process. 

A promising device for inspecting difficult to reach spaces inside machines, structures, and organisms are vine robots; a soft, inflatable robot that grows from its tip like a plant \cite{Hawkes2017}.
The device is composed of a long thin tube of airtight film or fabric, inverted inside itself such that when pressurized it everts and extends from its tip.
Unlike a borescope or endoscope which is pushed from its base and can only traverse limited paths, the vine robot experiences no relative movement between its skin and the surrounding allowing it to navigate tortuous, cluttered paths with multiple curvatures.
These robots have been developed to less-invasively explore archaeological ruins \cite{Coad2020}, collapsed buildings \cite{Maur2021roboa}, and endovascular surgery \cite{Li2021}.

However, current vine robot designs are limited by how much force they can exert, and how tight of bends they can make. 
Most prior steering designs work by actively shortening one side of the robot so that it curves towards that side.
The most common method is to use pneumatic artificial muscles \cite{Coad2020, Greer2019, Kubler2023, Abrar2021, Naclerio2018}, however these designs have limited curvature and stiffness.
Similarly, pull tendons have been used for steering \cite{Naclerio2020, Wang2020, Blumenschein2021}, but are limited by friction and implementation complexity.
Another method is to use a rigid device inside the vine robot that can create tight curves and discrete locations \cite{Haggerty2021, Maur2021roboa, Satake2022}, at the expense of adding a rigid component to an otherwise soft robot. 
Further, various selective stiffening methods of varying complexity have been added to these designs to selectively control which parts of the robot bend \cite{Wang2020, Do2020, Do2023}.

An inherent limit to steering a vine robot by contracting one side is that the contractile method is antagonistic to the body pressure of the robot, reducing how much force it can exert on the environment.
Most studies of vine robot have focused on how little force they exert \cite{Haggerty2019}, which would be beneficial for navigating delicate surgery environments, but perhaps less useful for inspection tasks that may require opening a door or turning a lever.
Although the robot can exert an axial force at its tip \cite{Haggerty2021}, they have poor bending stiffness \cite{Satake2022} without an active stiffening mechanism \cite{Do2023}.

\begin{figure}
  \begin{center}
  \includegraphics[width=\columnwidth]{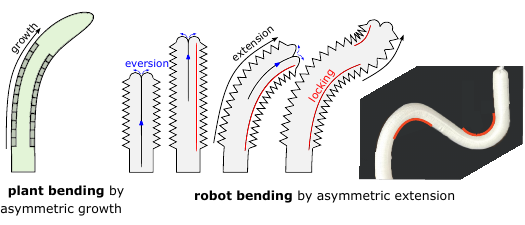}
  \caption{Like a plant, the vine robot grows and bends by asymmetrically extending one side of its body. The keys to this design are an anisotropic skin that allows the robot to extend axially, and layer jamming locking bodies along its inside that prevent one side from extending to create bends. For scale, robot is 32 mm in diameter.
  }
  \label{fig:KeyFeatures}
  \end{center}
  \vspace{-3mm}
\end{figure}

An alternative steering method inspired by the biological growth of plants is to lengthen one side of the robot and create a bend in the opposite direction. 
Plants do not have contractile muscles, instead they selectively elongate the cells on one side of their body and rely on their internal turgor pressure to bend \cite{burgert2009actuation}. 
This steering method would be advantageous for a vine robot because its steering method and internal pressure do not work counter to each other, allowing the robot to maintain both high body stiffness and high curvature at the same time. 
One complicated design using selective release of latches was implemented in a vine robot \cite{Hawkes2017}, but its curves were permanent.

In this work we present a method of reversible vine robot steering inspired by biological growth that allows for higher forces and tighter curvatures than existing designs. 
To do this we developed a highly-anisotropic, wrinkled, composite film that allows the robot to extend over 120\% axially, but only 3\% circumferentially. 
The robot selectively steers by stiffening one side of the robot with an internal jamming structure driven by the body pressure of the robot, and allowing the other side of the robot to extend (Fig. \ref{fig:KeyFeatures}).
The key advantage of this design over other steering and jamming designs is that the body pressure of the robot both pressurizes the jamming structure and drives bending, allowing it to reversibly exhibit high curvature and exert high forces at the same time. 
This could be useful when inspecting difficult to reach spaces such as inside an aircraft engine. 
What follows is a more detailed description of the design (Sec. \ref{sec:Design}), basic analytical models of its performance (Sec. \ref{sec:model}), robot demonstrations and characterizations of its performance (Sec. \ref{sec:Results}), and concluding thoughts (Sec. \ref{sec:Discussion}).



\section{Design}
\label{sec:Design}
\begin{figure}
    \centering
    \includegraphics[width=\columnwidth]{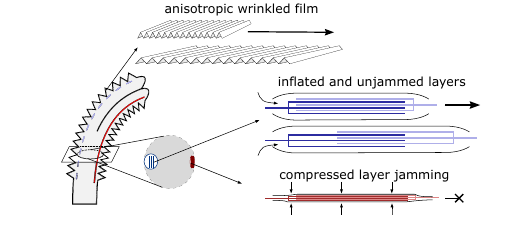}
    \caption{The robot bends by letting its uniaxially-wrinkled composite film stretch while one side of the robot is locked by a layer jamming locking body compressed by the internal body pressure of the robot. }
    \label{fig:design}
\end{figure}

In this section, we describe the inspiration for the design from the mechanisms of plant bending as well as the two key components that enable the device--the anisotropic composite film and the integrated layer jamming structures (Fig. \ref{fig:design}).

\subsection{Plant-Inspired Bending by Lengthening}
Bending in plants is a complicated process mediated by various hormones. 
However, at a high level, the principles involved can serve as guides for bio-inspired robotic design. We briefly describe the mechanism of plant bending below, based on \cite{burgert2009actuation,jonsson2023multiple}.

Without contractile muscles, plants generally rely on extension to bend. 
Extension requires lengthening of the stiff cell walls that work for the structure of plants (Fig. \ref{fig:KeyFeatures}). 
These cell walls are composed primarily of load-bearing cellulose fibers cross-linked with other organic molecules including hemicellulose and pectin.
To bend, the hormone expansin cleaves the crosslinks, allowing the cellulose fibers to slide past one another in the axial direction; this motion is driven by the high internal pressure of the plant. However, the circumferential structure is maintained, such that the plant does not swell radially.


The key principles that can be gleaned from this description to create plant-inspired bending are two-fold.
First, a pressurized, tube-like structure will lengthen axially but not swell radially if it is formed from a highly anisotropic skin that is stiff in the circumferential direction. 
Second, this structure will bend if the lengthening is controlled to be different on different sides; varying the interlocking among overlapping, inextensible fibers controls lengthening.
Next we describe our engineered solutions based on these principles.

\subsection{Anisotropic Composite Film}
To achieve bending by elongation, we need an air-tight, highly anisotropic film with a high specific strength.
The skin needs to be on the order of 20x stiffer in one direction than the other to achieve the desired steering curvature (see Sec. \ref{sec:model} \eqref{eqn:R}). 
To meet these requirements, we developed a composite material comprising one layer of a uniaxially pre-stretched elastomer laminated to an inelastic film.
When the pre-stretch of the elastomer layer is relaxed, the inelastic film spontaneously buckles to form a uniformly wrinkled surface. 
The film is then formed in a tube, with the wrinkles running around the circumference of the tube.
Such spontaneous buckling phenomenon in thin films has been described \cite{jiang2007finite,huang2005nonlinear,bowden1998spontaneous}, and explored for stretchable electronics \cite{lipomi2011stretchable}, but not as a structural, anisotropic material.

\subsection{Integrated Layer Jamming Structures}
To control the lengthening of different sides of an extensible tube, we need structures that can controllably lengthen. Inspired by the controllable interlocking among cellulose fibers, we use pressure-driven layer jamming structures attached along the length of the tube.
The jamming structure is composed of several interlocking sheets \cite{choi2017soft, kim2013novel} of high-friction material inside a thin flexible tube.
When the tube is compressed by an internal vacuum or external pressure, the layers jam against each other and can no longer easily slide past one another.
When the tube is inflated, the layers are free to slide past each other. 
We note that various soft robotic systems have utilized jamming technologies\cite{fitzgerald2020review}, including one example using layer jamming structures to stiffen and create virtual joints for dynamic reconfiguration in vine robots \cite{Do2023}. 

\subsection{Implementing Bending by Lengthening}
Our robot is composed of a tube of our anisotropic composite film with several independent jamming structures attached along the length of the inside of the main body.
The main body is inverted, such that when pressurized, it will evert and lengthen from its tip.
The body pressure compresses the jamming structures, so that they are in a default jammed state.
The jamming structures can be independently inflated to release them and cause a bend.
Note that the jamming structures are arranged in series along each side of the robot to control shape. 
More jamming structures allow for more controllable degrees of freedom in the robot.
Importantly, very little air needs to flow in and out of the jamming structures to control them, such that very small tubes can be used to control each (see Sec. \ref{sec:fab:assem}).
This contrasts with designs that use artificial muscles to control shape, wherein large volumes of air must flow in and out of the muscle, requiring large tubing.
Instead, in the proposed design, the work of turning is done with the main body, freeing the steering mechanisms to do no work, essentially becoming brakes instead of actuators.

\section{Modeling}
\label{sec:model}

In this section, we describe simple mathematical models of the proposed device that serve as tools for design.


\subsection{Geometric Kinematics}
\label{sec:model:geometry}
\begin{figure}[b]
  \begin{center}
  \includegraphics[width=\columnwidth]{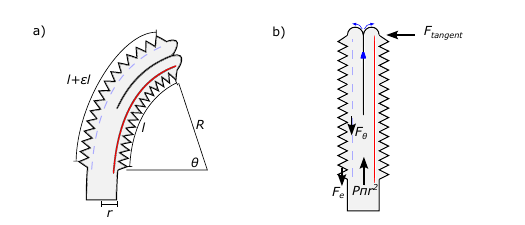}
  \caption{Model geometry (a) and forces (b).
  }
  \label{fig:ModelParameters}
  \end{center}
  \vspace{0mm}
\end{figure}

The first model describes how geometry and maximum strain relate to the maximum curvature of the robot body. For a given strain $\varepsilon$ on one side of the robot body, we can model the radius of curvature $R$ of the resultant bend by looking at arc lengths $l$ and $l+\varepsilon l$ (Fig. \ref{fig:ModelParameters}a).
\begin{equation}
    l=R \theta
\end{equation}
and 
\begin{equation}
    l+\varepsilon l = (R+2r)\theta
\end{equation}
where $R$ is the radius of curvature, $r$ is the radius of the robot, and $\theta$ is the angle of curvature.
Solving for $R$ and $\theta$ we find
\begin{equation}
    R=2r/\varepsilon
    \label{eqn:R}
\end{equation}
and
\begin{equation}
    \theta=\varepsilon l/(2 r).
    \label{eqn:theta}
\end{equation}

This model helps inform the amount of stretch required in the anisotropic film if the radial swelling and desired radius of curvature are set by the designer. We set the radial swelling to be less than 5\% to approximately maintain the robot radius during growth, and we aim for a radius of curvature equal to the robot diameter to access highly constrained environments. In this case, the required axial strain is 100\%, and the material should be at least 20x stiffer in the circumferential direction than the axial direction. 

\subsection{Force Curvature Equilibrium}
\label{sec:model:curve}
The second model explores how as the robot is pressurized, its shape changes; this can be calculated using statics with force and moment balances.
We assume that the internal forces acting on the robot are internal pressure $P$ times area $\pi r^2$, an elastic force 
\begin{equation}
    F_e=EA_{film}\varepsilon l
\end{equation}
from the axial elasticity of the body where $E$ is the film's modulus and $A_{film}$ is the cross sectional area of the film, and an angle dependent friction force 
\begin{equation}
    F_{\theta}=K_{\theta}e^{\mu \theta}
\end{equation}
from capstan friction due to internal resistance in a deactivated jamming structure where $K_{\theta}$, and $\mu$ are empirical constants. 

If no jamming structure is activated, then the length of the robot $l$ is simply 
\begin{equation}
    l = P\pi r^2/EA_{film}+l_0
\end{equation}
where $l_0$ is its depressurized length.

If one jamming structure is activated, we can do a moment balance around that side of the body, resulting in 
\begin{equation}
    P\pi r^3 = (EA_{film}\varepsilon l + K_{\theta}e^{\mu \theta})2r
    \label{eq:moment_balance}
\end{equation}
for which there is no explicit solution for $\varepsilon$, $R$, and $\theta$.

This model helps inform the design of the robot. First, $E$ should be minimized to reduce the required pressure for a given strain. Second, it informs how large of a force the jamming structure on the inner curve of the robot must be able to withstand.
We note that the axial force must be equal in the inner wall to that in the outer wall for static equilibrium, and thus from \eqref{eq:moment_balance}, we know that the force that the jamming structure must withstand is $EA_{film}\varepsilon + K_{\theta}e^{\mu \theta}$.

\subsection{External Force Application}
The third model describes how much force the vine robot can apply in the direction perpendicular to its long axis by bending. 
When locking one side of the body, the robot can apply a perpendicular force $F_{tangent}$ at its tip as it is pressurized.
This can be solved by a moment balance as
\begin{equation}
    F_{tangent}=(P\pi r^3-E\varepsilon-K_\theta e^{\mu\theta})/l.
\end{equation}
For comparison, the force that a robot with a pneumatic artificial muscle with force $F_m(\varepsilon) = (\pi r_m)P_m[a(1-\varepsilon)^2-b]$ modeled by the ideal McKibben muscle equation \cite{Naclerio2018,tondu2012modelling} would be
\begin{equation}
    F_{tangent}=(2r*(\pi r_m)P_m[a(1-\varepsilon)^2-b] - P\pi r^3)/l.
\end{equation}

Notice that in the proposed device, the force is directly related to the pressure inside the body, meaning more force can be attained by increasing the body pressure.
In contrast, for the previous design that uses an artificial muscle, the body pressure is actually working against the muscle pressure, meaning that a stiffer, higher pressure body inhibits external force application.



\section{Fabrication}
\label{sec:Fabrication}

\subsection{Anisotropic Composite Film}
\label{sec:fab:film}
Our composite anisotropic robot skin is a laminate of a pre-stretched membrane and an inelastic film and fabricated as follows.
First, a rectangular piece of 52 $\mu$m thick thermoplastic polyurethane (TPU) is uniaxially pre-stretched up to 200\% and fixed at its ends.
Second, a rectangular piece of 42 $\mu$m thick Dyneema composite fabric (ultra-high-molecular-weight polyethylene fibers sandwiched between polyseter film) is cut to the length of the pre-stretched TPU and coated with 65 $\mu$m thick pressure sensitive adhesive (PSA) transfer tape (3M F9460PC).
Finally, the two layers are adhered together and allowed to contract, forming wrinkles.


\begin{figure}[t]
  \begin{center}
  \includegraphics[width=\columnwidth]{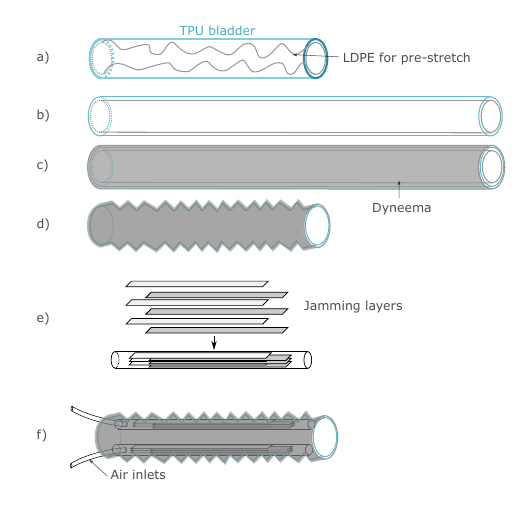}
  \caption{Fabrication of the robot. (a, b) The TPU bladder is pre-stretched by an LDPE tube. (c) Dyneema composite fabric as attached to the TPU. (d) The LDPE tube is deflated and removed, leaving a wrinkled robot body. (e) The locking body is assembled of two sets of alternating strips of plastic and placed inside a TPU tube. (f) The locking bodies are attached to the inside of the robot body.
  }
  \label{fig:Fabrication}
  \end{center}
  \vspace{0mm}
\end{figure}

\subsection{Anisotropic Stretchable Tip-everting Body}
\label{sec:fab:tube}
The body of the robot is composed of the anisotropic film described in Section \ref{sec:fab:film}, but formed in a tube with a different method shown in Fig. \ref{fig:Fabrication}.
First a 30 cm long, 74 mm diameter bladder of 52 $\mu$m thick TPU is formed by heat sealing.
Second, the TPU bladder is placed around a scrunched-up, double-walled, 90 cm long, 32 mm diameter tube of 50 $\mu$m thick low density polyurethane (LDPE) tube.
Third, the LDPE tube is inflated to 40 kPa to pre-stretch the TPU bladder by 200\%.
Fourth, the pre-stretched TPU tube is coated in 65 $\mu$m thick PSA tape and a sheet of 42 $\mu$m thick Dyneema composite fabric.
Finally, the LDPE tube is deflated and removed, leaving behind an axially wrinkled, 32 mm diameter anisotropic composite film tube.

\subsection{Layer Jamming Locking Body}
\label{sec:fab:jamming}
The locking body is composed of two sheaves of interlocking film strips inside a flexible bladder as shown in Fig. \ref{fig:Fabrication}e.
The two sheaves are made of five and six 5 mm wide, 170mm long, and 0.127 mm thick Duralar sheets respectively, each joined together at one end.
The strips are interwoven for a length of 150 mm, placed inside a heat-sealed TPU bladder, and secured at each end to the TPU tube by PSA tape.
The sheaves can slide past each other unless the TPU bladder is compressed by external pressure.
\subsection{Assembly}
\label{sec:fab:assem}
The locking bodies are attached to the inside of the robot body.
To do this we invert and stretch the robot body then attach the stretched locking bodies along its length with PSA tape. 
Next we add 1.2 mm diameter silicone air tubes to the locking bodies to provide positive pressure to un-jam them.
Finally, the robot body is inverted again for the final configuration shown in Fig. \ref{fig:Fabrication}f.

Multiple locking bodies can be attached in series along the length of the robot body to enable compound curvatures.


\section{Results}
\label{sec:Results}

\begin{figure}[b]
  \begin{center}
  \includegraphics[width=\columnwidth]{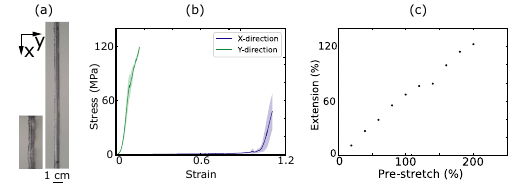}
  \caption{Stress-strain curves of the anisotropic material in the longitudinal and transverse direction. (a, b) Longitudinal corresponds to the direction with presence of wrinkles and transverse direction is its orthogonal direction that lacks wrinkles. (c) Data showing that pre-stretch in the fabrication increased the ability of the anisotropic material to stretch in the longitudinal direction.
  }
  \label{fig:skin_characterization}
  \end{center}
  \vspace{0mm}
\end{figure}

\subsection{Component Characterization}
In this section, we describe characterization tests of the two key components of the system: the anisotropic skin material and the jamming structures. 

\subsubsection{Anisotropic Skin Characterization}
\label{sec:ModelSubsection 1}

To characterize our anisotropic skin, we measured its elastic moduli through uniaxial testing on an Instron machine at a rate of 20mm/min (five rectangular specimens, 6 mm x 50 mm for both X and Y directions). Fig. \ref{fig:skin_characterization}a shows the material before and after stretch in the X direction (along robot body) and Fig. \ref{fig:skin_characterization}b shows the stress-strain curves.
The results in the X direction indicate that the behavior of the material is composed of two regimes--one low stiffness during unwrinkling and one high stiffness once the wrinkles are taut.
In contrast, the stress-strain curve in the Y direction presents one high-stiffness regime due to the lack of wrinkles in this direction of the anisotropic material.
Importantly, we show a high anisotropic ratio: the elastic modulus in the X direction is 1.98 MPa, and in the Y direction, 879.1 MPa, representing a ratio of over 400x. This exceeds the specification described in Sec. \ref{sec:model:geometry}. 


To characterize the effect of pre-stretch of the TPU film during fabrication on the stretchability of the anisotropic material in the X direction, we fabricated specimens with varied pre-stretch. We measured the percent extension $e = (L_{max}-L_0)/L_0$, where $L_{max}$ is the max length and $L_0$ is the initial length (Fig.  \ref{fig:skin_characterization}c). Increasing the pre-stretch for each specimen increased the ability to stretch nearly linearly.

\subsubsection{Jamming Structure Characterization}
\label{sec:ModelSubsection 6}

\begin{figure}[t]
  \begin{center}
  \includegraphics[width=0.7\columnwidth]{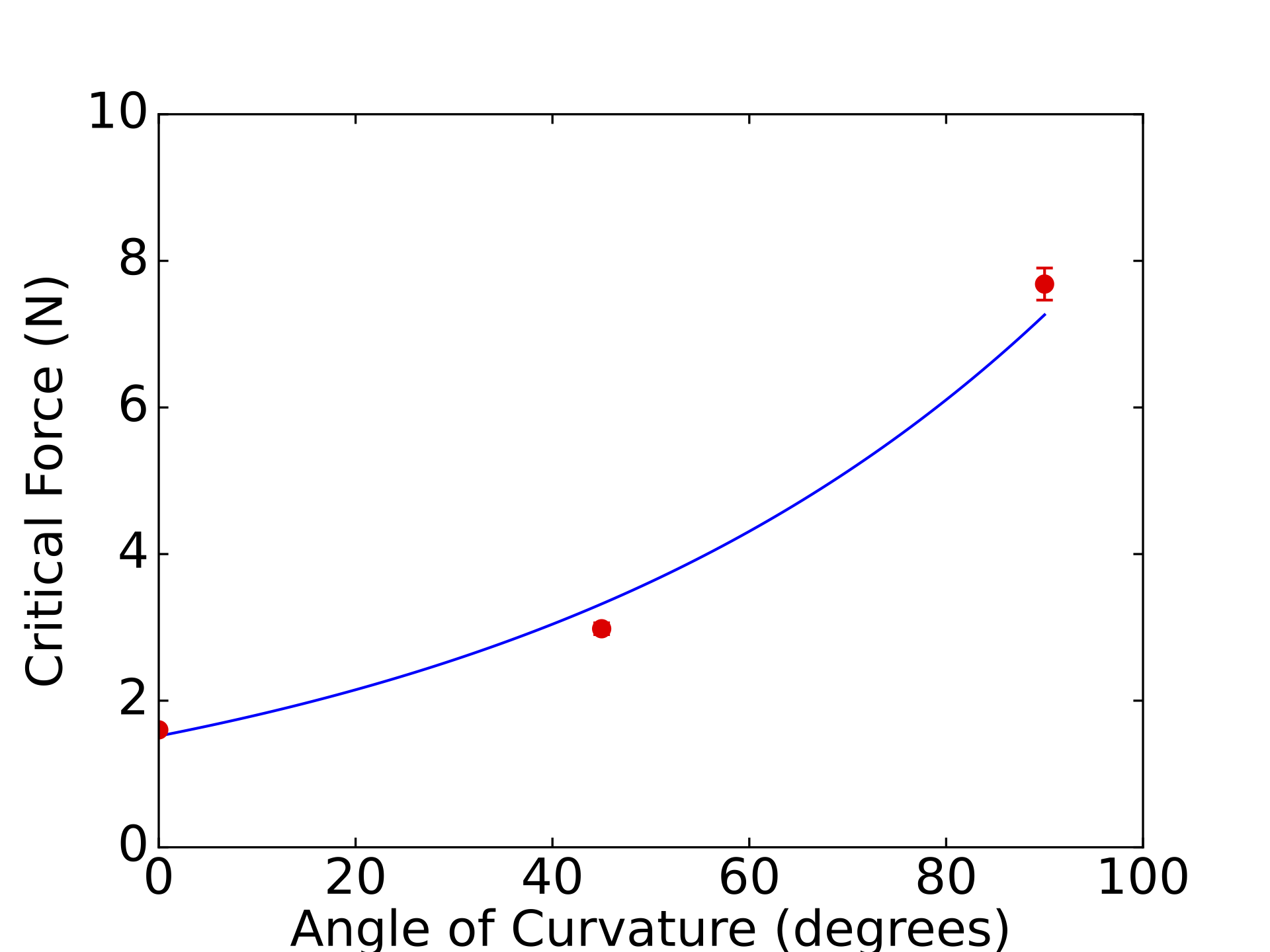}
  \caption{Increasing the curvature of the jamming structure, even when no vacuum pressure is applied, causes an increase in the critical tension force required to cause a slip. An exponential fit is plotted, suggested by the model from \eqref{eq:moment_balance}. 
  }
  \label{fig:FcVpressure}
  \end{center}
  \vspace{0mm}
\end{figure}

We built one jamming structure to characterize the effect of the curvature on the critical force before slipping. The layer size for jamming unit was 5x90 mm with an overlap of layers of 30 mm. 
We applied tension to the jamming structure while held over varied 3D-printed arcs to change the angle of curvature, while maintaining 30 mm in contact with each arc (n=5 for each arc).
As shown in Fig. \ref{fig:FcVpressure}, the critical force increased exponentially as the deflection angle increased. Such trend suggests that the critical force in the inner side of the curve of the vine robot is higher than when the jamming unit is straight. 

\subsection{Straight Vine Body Characterization}
\label{sec:ModelSubsection 2}

\begin{figure}[b]
  \begin{center}
  \includegraphics[width=1\columnwidth]{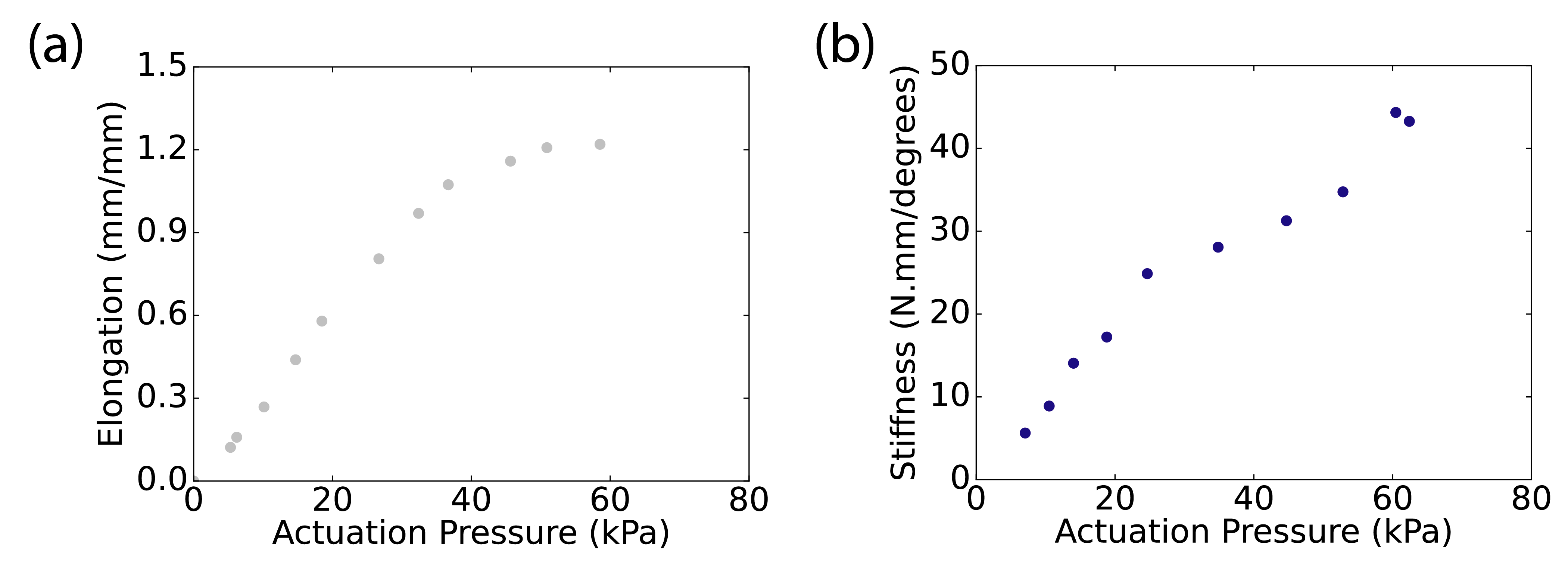}
  \caption{(a) Elongation of vine body versus the actuation pressure. (b) Normalized stiffness-pressure curve of the vine body prior to attaching the jamming structures.
  }
  \label{fig:ExtensionVPressure}
  \end{center}
  \vspace{0mm}
\end{figure}
Since the main body of our vine robot exhibits a new working principle based on its ability to stretch in the longitudinal direction, it is important to characterize the effect of actuation pressure lengthening. 
We inflated the robot body in a range from 0 kPa to 60 kPa and measured the length at each increment. 
Note that no jamming structures were attached.
Fig. \ref{fig:ExtensionVPressure} shows that increasing the actuation pressure increased the length linearly at first, before saturating as the wrinkles became stretched taut.  
The vine body was able to stretch two times its original length, such elongation is slightly higher in a flat sheet with the same pre-stretch due to differences between stretching a flat sheet and an airtight tube (Fig. \ref{fig:skin_characterization}).

Second, we characterized the bending stiffness of the vine body in terms of normalized torque (N.m/degrees) at different pressure values. 
Ten sets of actuation pressures were used to inflate the vine body. To obtain the stiffness for each pressure, we increased the lateral load at a set length, recording the vertical displacement. The stiffness was calculated using the slope of the torque-deflection angle curve and the known set length.  
Increasing the pressure increased the normalized stiffness of the inflatable beam as expected, shown in Fig. \ref{fig:ExtensionVPressure}b. 

\subsection{Free Strain for Steering}
\label{sec:ModelSubsection 3}

\begin{figure}[tb]
  \centering
  \includegraphics[width=0.9\columnwidth]{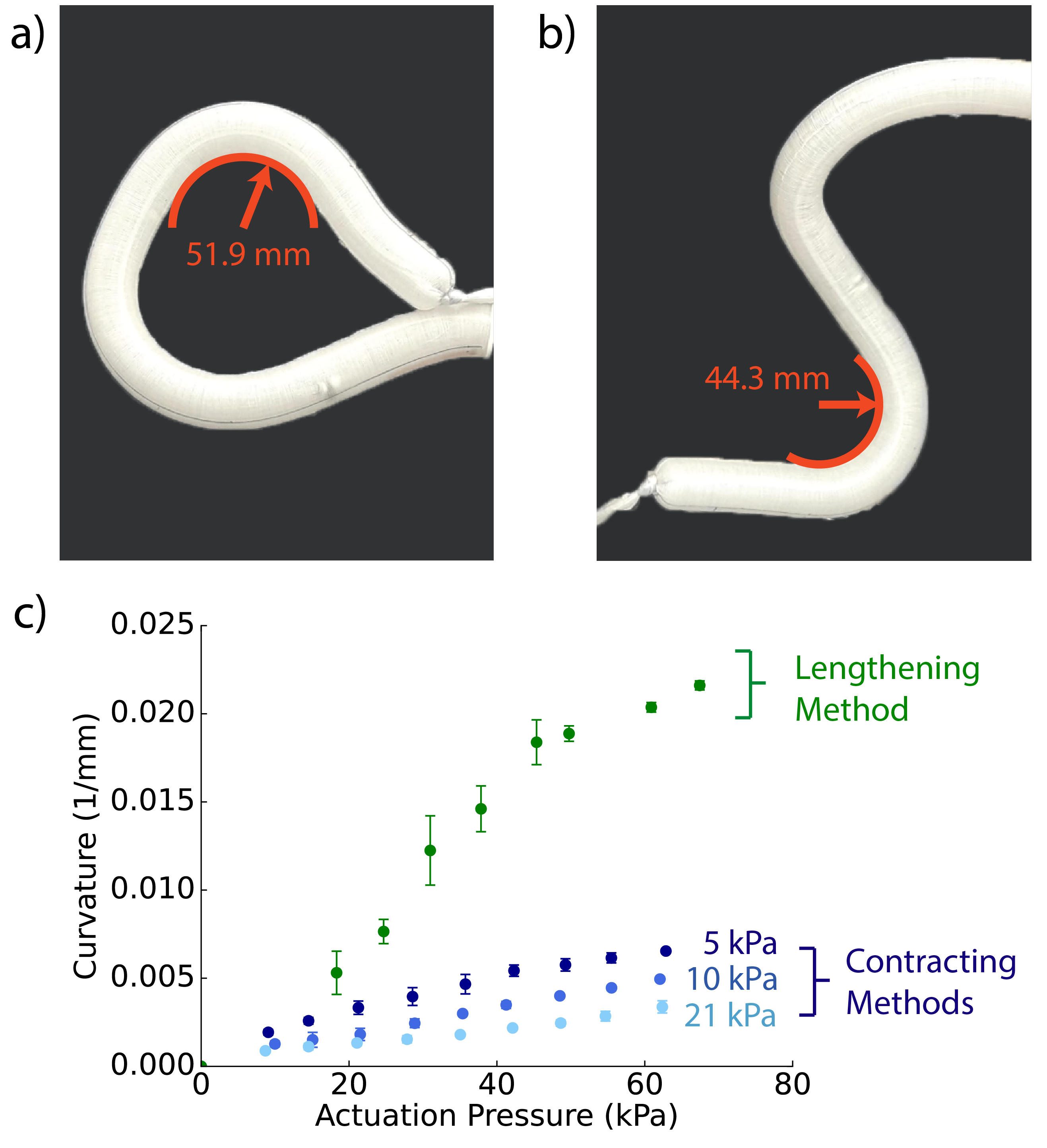}
  \caption{(a, b) The robot can create tight, compound curvatures with a radius of curvature as tight as 1.4x its body diameter. (c) Curvature of the vine robot vs. actuation pressure is much higher with our lengthening method (top curve) compared to prior contracting methods using an artificial muscle. Tests with the artificial muscle were performed for three different robot body pressures (lower three curves).}
  \label{fig:RhoVactPressure}
  \vspace{0mm}
\end{figure}

For navigation tasks without substantial forces resisting bending, characterizing bending radius in free space is important.
First, we formed the robot into to tight curvature shapes, showing a full circle and a tight ``S'' (Fig. \ref{fig:RhoVactPressure}, a-b). We found radius of curvatures of 44 to 52 mm, allowing the robot to turn around a radius less than two body diameters.
Second, we characterized our method of steering against gravity and compared it to a control using a previous vine robot steering method, external artificial muscles, fPAMs \cite{Naclerio2020}.
To characterize this behavior, the vine robot was set horizontally on a table and allowed to curve up away from the table.
We recorded the radius of curvature of the vine robots using a video as the actuation pressure was increased from 0 to 60 kPa.
For our bending by lengthening vine robot, the jamming structure away from the table was inflated and relaxed.
For the bending by shortening vine robot, the actuation pressure was the applied pressure in the artificial muscle, and the body pressure was varied.
Fig. \ref{fig:RhoVactPressure}c shows that for all cases of internal pressure for the control robot, our method of steering by lengthening exhibits greater than 3x and 6x more curvature.

\subsection{Bending to Apply External Force}
\label{sec:ModelSubsection 4}

\begin{figure}[t]
  \begin{center}
  \includegraphics[width=1\columnwidth]{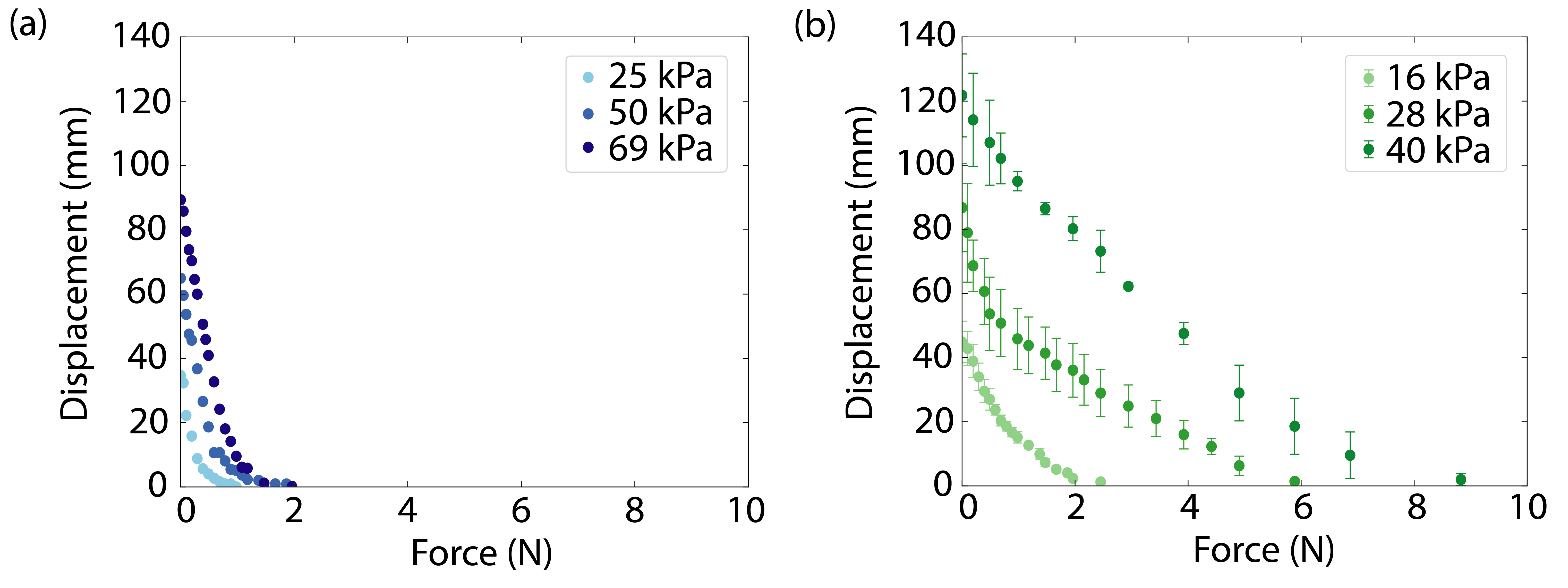}
  \caption{The force vs. displacement curve of a cantilevered vine robot using an artificial muscle to shorten one side (a) or our method to lengthen one side (b) as a force is applied 19 cm from its base. The displacement is a measure of how far off the table the robot could lift the force. Both the force and displacement of our robot that bends by lengthening exceeds that of a same-sized robot that bends by shortening.
  }
  \label{fig:LiftForceVdispl}
  \end{center}
  \vspace{0mm}
\end{figure}

We hypothesize that our steering method can exert substantial force since the internal pressure is causing the bending, rather than resisting it, as in the case of previous work with artificial muscles.
To test this we compared how high a cantilevered robot with our method or a pneumatic muscle (fPAM \cite{Naclerio2020}) could lift masses at a position of 19 cm from its base.
The force vs. deflection curve was measured for three different actuator pressures shown in Fig. \ref{fig:LiftForceVdispl}, with three trials (n = 3) each.
The body pressure of the robot with the fPAM was kept constant at 10 kPa.


The displacement-load curves for with fPAM and our robot are shown in Fig. \ref{fig:LiftForceVdispl}a and b respectively.
These results indicate that fPAMs could not exert meaningful displacement with forces above 1 N.
Conversely, our method could exert forces up to 10 N. 
This suggests that steering by lengthening is better at exerting forces on the environment while steering than can be done with artificial muscles.

\subsection{Demonstrations}
\begin{figure}
  \begin{center}
  \includegraphics[width=\columnwidth]{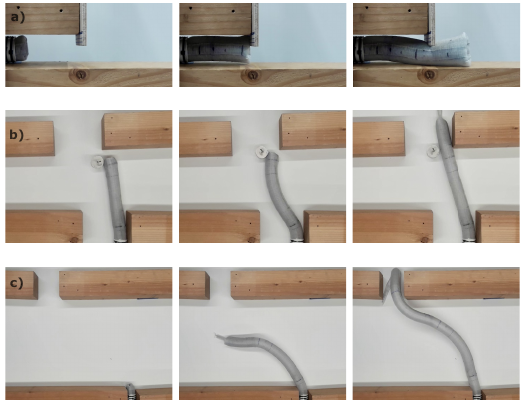}
  \caption{(a) Robot fits through a tight gap. (b) Robot pushes a mass out of the way to reach the goal.
  (c) Robot creates a compound curvature to reach the goal. For scale, robot is 32 mm in diameter.
}
  \label{fig:Demo}
  \end{center}
  \vspace{0mm}
\end{figure}

We performed three demonstrations of the robot's ability to traverse obstacles that could be found in an inspection task, as shown in Fig. \ref{fig:Demo}.
First (Fig. \ref{fig:Demo}a), the 32 mm robot squeezed through a 25 mm gap, something that a rigid robot could not do.
Second (Fig. \ref{fig:Demo}b), the robot pushed a 200 g weight out of the way to access its target.
And finally (Fig. \ref{fig:Demo}c), the robot grew while making a compound curvature to reach an opening oriented parallel, but offset from its starting point.

\section{Discussion}
\label{sec:Discussion}
One limitation of our proposed anisotropic material is that it has to be fabricated by hand from two existing materials.
Future work could explore manufacturing methods to create the material via scalable methods.
A limitation of the proposed robot as built is its length. Future versions should be extended to increase the usefulness of the robot. 
Ideally, as the length grows, the number of layer jamming sections will increase to increase the degrees of freedom of the device. 
Although this will add more tubing lines, as noted earlier, these lines can be very small (1.2mm diameter), such that many can be added without hindering robot performance. This contrasts with lines to pneumatic actuators, which need to have a larger diameter (usually around 4-6 mm) to allow enough flow, limiting the feasible number of degrees of freedom compared to the proposed design.


\section{Conclusion}
We have presented an alternative way of steering a vine robot by lengthening one side using an aniostropic body material and layer jamming.
This plant-inspired method exhibits higher forces and curvatures than methods that work by shortening with artificial muscles. 
This is an important advancement toward the development of practical vine robots for exploring and accessing difficult to reach spaces that no other device can reach, such as inside machinery or through tortuous lumens in the human body. 
Additionally, the composite material offers new properties for the field of soft robotics generally.





\bibliographystyle{IEEEtran}
\bibliography{IEEEabrv,Bibliography}

\end{document}